\documentclass[10pt,journal,letterpaper,compsoc]{IEEEtran}
\usepackage{cite}
\usepackage{amsmath,amssymb,amsfonts}
\usepackage{algorithmic}
\usepackage{graphicx, subfigure}
\usepackage{textcomp}
\usepackage[export]{adjustbox}
\begin{document}
%
%
\title
  {
    Adversarial Reinforcement Learning Framework for Benchmarking Collision Avoidance Mechanisms in Autonomous Vehicles\thanks{This work was supported by the National Science Foundation (NSF) (NSF-CRII-CPS-1743490). Any opinions, findings, and conclusions or recommendations expressed in this material are those of the author and do not necessarily reflect the views of the NSF.}
  }
%
%
\author
  {
		\IEEEauthorblockN
    {
    Vahid Behzadan\IEEEauthorrefmark{1} and
    Arslan Munir\IEEEauthorrefmark{3}
    }

    \IEEEauthorblockA
    {
    Email:                                 %
    \IEEEauthorrefmark{1}behzadan@ksu.edu,
    \IEEEauthorrefmark{3}amunir@ksu.edu
    }
  }

%
%
\IEEEcompsoctitleabstractindextext{
\begin{abstract}
With the rapidly growing interest in autonomous navigation, the body of research on motion planning and collision avoidance techniques has enjoyed an accelerating rate of novel proposals and developments. However, the complexity of new techniques and their safety requirements render the bulk of current benchmarking frameworks inadequate, thus leaving the need for efficient comparison techniques unanswered. This work proposes a novel framework based on deep reinforcement learning for benchmarking the behavior of collision avoidance mechanisms under the worst-case scenario of dealing with an optimal adversarial agent, trained to drive the system into unsafe states. We describe the architecture and flow of this framework as a benchmarking solution, and demonstrate its efficacy via a practical case study of comparing the reliability of two collision avoidance mechanisms in response to intentional collision attempts.

\end{abstract}
}
\maketitle
\IEEEdisplaynotcompsoctitleabstractindextext
\IEEEpeerreviewmaketitle
\sloppy

%
%
\section{Introduction}
\label{sec:introduction}
It is widely believed that the transportation systems of future will be dominated by autonomous vehicles (AVs). With the rapid advancements of this field in recent years, many have come to predict that this shift will occur within the next ten years. A major motivation for the interest and push towards development of AVs stems from the demand for safer transportation. It is generally assumed that replacing the intrinsic imperfections of human drivers with expert computational models may significantly reduce the number of accidents caused by driver error \cite{bagloee2016autonomous}. Yet, development of reliable and robust AV technologies remains an ongoing challenge, and is actively pursued from various directions of research and development \cite{koopman2017autonomous}.

Of particular importance is the research on reliable motion planning and collision avoidance mechanisms. Over the span of multiple decades, numerous approaches towards this problem have been proposed~\cite{mohanan2018survey}, ranging from control theoretic formalizations and optimal control methods to potential field- and rule-based techniques. More recently, advances in machine learning have enabled new data-driven approaches to collision avoidance based on techniques such as imitation learning \cite{hussein2017imitation} and deep Reinforcement Learning (RL)\cite{arulkumaran2017brief}. However, with the growing complexity in their deployment settings and mechanisms, the challenge of providing safety guarantees on these solutions is becoming increasingly difficult \cite{koopman2017autonomous}. A notable instance is the Traffic Collision Avoidance System (TCAS), which had satisfied the rigorous safety requirements of the Federal Aviation Authority (FAA) before its wide deployment in NextGen commercial aircraft. Yet, it was recently shown \cite{jun2014causal} to be highly unreliable in the modern high-density airspace, to the extent that it may give rise to Inevitable Collision States (ICS)---for which, regardless of the future trajectories, a collision eventually occurs. Furthermore, recent research demonstrate that automatic sense and avoid mechanisms can be adversarially exploited to manipulate the motion trajectory of AVs \cite{pierpaoli2015altering}. 

In response, a growing number of mitigation techniques and novel approaches to safe motion planning are proposed, but each with certain case-specific assumptions and ad hoc verification procedures. Consequently, quantitative comparison across such methods is rendered extremely difficult. Current state of the art includes several attempts towards benchmarking of safe behavior in motion planning and collision avoidance (e.g., \cite{moll2015benchmarking, calisi2008unified}), but many of the current frameworks fail to meet the requirements of new adaptive techniques based on machine learning \cite{koopman2017autonomous}. Also, current benchmarking frameworks do not provide comprehensive and robust mechanisms for exploration in the complex spaces of undesired states and trajectories. Prominent approaches in such frameworks are based on randomized or scenario-based generation of obstacles (e.g., \cite{calisi2008unified}), which are highly prone to missing critical ICS or other undesired states specific to the mechanism under test. Another approach in these frameworks relies on computationally expensive techniques for reachability analysis of collision states, which also fail to provide concrete guarantees on critical boundaries of safe operation \cite{zhou2018efficient}. 

Aiming to fill the gap in safety-focused benchmarking, this paper proposes a novel framework based on machine learning for benchmarking the reliability of novel techniques under the worst-case scenario of interacting with an optimal adversarial agent. This framework adopts the powerful exploration and optimization performance of deep RL to train adversarial autonomous agents whose goal is to learn optimal navigation policies that aim to drive the system into ICS and other unsafe states. Depending on the parameters and goals of analysis, such objectives may include direct collision of adversarial agent with the AV, or exploiting the collision avoidance mechanisms to manipulate the trajectory of AV to either alter and control its path, or to indirectly induce collisions between the AV and other objects in the environment. 

Building on this foundation, the main contributions of this work include:
\begin{enumerate}
	\item Proposal of a computational framework and process flow for worst-case benchmarking of collision avoidance algorithms, independent of their complexity, stochasticity, and even adaptive dynamics;
	\item Proposal of a deep RL process flow to seamlessly adapt to the system under test, and to overcome the shortcomings of fully random or scenario-based exploration mechanisms; 
	\item Proposal of novel metrics for standardized comparison of collision avoidance algorithms;
	\item Demonstration of the practical application and efficacy of the proposed framework via a realistic case study of comparing the reliability of two collision avoidance mechanisms in response to intentional collision attempts.
	
\end{enumerate}

The remainder of this paper is organized as follows: Section \ref{sec-architecture} provides the architectural details of our proposed framework, and introduces novel metrics for quantification and comparison of motion planning and collision avoidance algorithms. Section \ref{sec-case} demonstrates the application of this framework through a practical case study of comparing the reliability of two collision avoidance mechanisms, viz., one based on a deep RL navigation policy, and another employing a control theoretics mechanism. Finally, Section \ref{sec-conclusion} concludes the paper with remarks on potential directions of further research.

\section{Proposed Architecture} \label{sec-architecture}
The high-level idea of this framework is to employ deep RL methods \cite{arulkumaran2017brief} for end-to-end training of an optimal adversarial policy (i.e., mapping of states to actions) for autonomous motion planning. The difference between this adversarial policy and typical AV policies is that the latter aims to achieve an optimal motion planning while avoiding collisions, while the former aims to learn an optimal motion planning with the aim of \emph{causing intentional direct or indirect collisions}. Through suitable choices of optimality criteria and the corresponding reward (i.e., objective) function, this approach enables the adoption of powerful state-space exploration and policy optimization techniques developed for deep RL applications. Furthermore, we argue that both the training and test-time procedures of adversarial policies provide quantitative measures of reliability, which can be used for benchmarking of behaviors in worst-case scenarios. Accordingly, our proposed framework is comprised of 4 main components: \emph{simulation environment, objectives, deep RL algorithm, and quantitative metrics}, detailed as follows:

\subsection{Simulation Environment}
Computational testing of safety in AV necessitates a simulation environment that takes into account the various aspects of deployment settings, including (but not limited to): the physics of the AV, the terrain and road conditions, number and type of other mobile or static objects in the environment, type of traffic, etc. Depending on the goals and criteria of the experiment, the simulation environment may be chosen from numerous open-source and commercial simulation platforms, such as TORCS \cite{loiacono2013simulated} and CARLA \cite{dosovitskiy2017carla}.

Aiming for a general-purpose solution, the proposed framework is designed to be independent of the simulation platform and setup. To this end, our proposed framework imposes the bare minimum requirements on the simulation environment, which are two-fold: First, the platform should provide an interface (i.e., API) to enable seamless integration with the deep RL module for external monitoring of state variables and control of agents; and second, the platform should allow for external adjustment of simulation flow and speed. The latter requirement is to eliminate the effect of training and computational lags on the flow of simulation.

If these requirements are met, the proposed framework can be seamlessly implemented on any configuration and setup of the environment, such as the choice of two-agent vs. multi-agent settings, selection of terrains, models of vehicular physics, and any other parameter configured for the experiment.

\subsection{Objectives}
\begin{figure} \label{fig:objective}
	\includegraphics[width=\columnwidth]{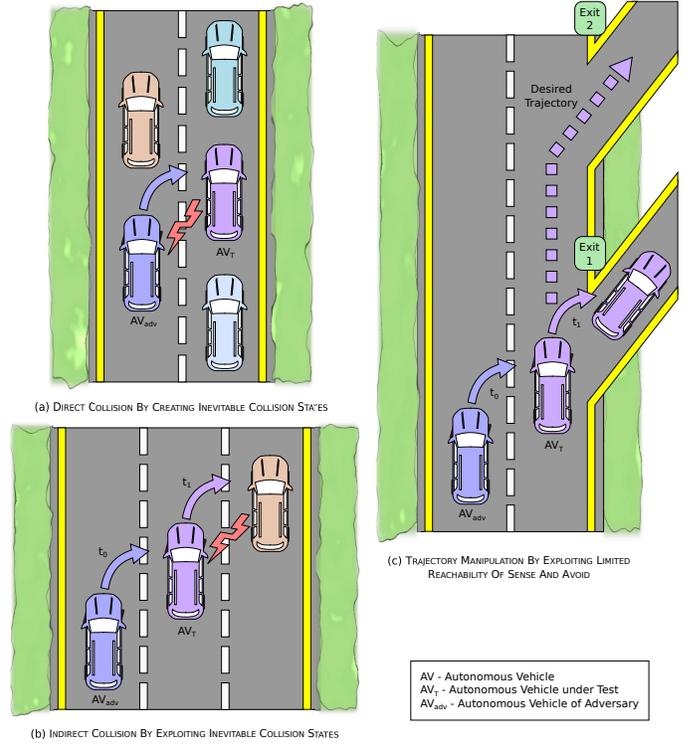}
	\caption{Illustration of adversarial objectives}
\end{figure}
Deep RL algorithms are based on optimization of action policies according to defined objectives, represented by a reward function $R(s_t, a_t)$, which produces numerical values corresponding to the instantaneous value of the action $a_t$ at state $s_t$. In the case of an AV aiming at traversing a race track in minimum time, a common formulation of the reward function is \cite{lillicrap2015continuous}:

\begin{equation}
R_t = V_x cos(\theta) - V_x sin(\theta) - V_x \vert d(t_p)\vert \label{Reward1}
\end{equation}

where $V_x$ is the velocity of AV along its longitudinal axis, $\theta$ is the angle between direction of AV and the track axis, and $d(t_p)$ is the distance between AV and the center of the track axis. This function rewards trajectories with high velocities in the direction of the track, but it does not account for the negative effect of collisions. A simple remedy is to add a negative reward term to $R_t$ when a collision is detected, for instance:

\begin{equation}
R_t = V_x cos(\theta) - V_x sin(\theta) - V_x \vert d(t_p)\vert - \eta\times C \label{Reward2}
\end{equation}
where $\eta \in \{0, 1\}$ is a binary variable representing whether a collision has occurred or not, and $C > 0$ is the negative reward whose magnitude determines how undesired collisions are.

As depicted in figure \ref{fig:objective}, experiments on an AV-under-test (henceforth referred to as $AV_T$) can employ one of the three choices as the high-level objective of the adversarial RL agent (designated by $AV_{adv}$), described below:
\begin{enumerate}
	\item \emph{Direct collision with $AV_T$}: This objective aims to find an optimal navigation policy that leads to direct collision of $AV_{adv}$ with $AV_T$. Optimality can be defined arbitrarily (e.g., minimum time to cause collision). An instance of corresponding adversarial reward functions is $R_t = \eta\times C' - d(AV_T, AV_{adv})$, where $d (.,.)$ is an arbitrary distance metric, and $C'$ represents the cost of direct collisions for $AV_{adv}$.
	\item \emph{Induced collision}: Unlike the former case, this objective aims to find an optimal navigation policy that manipulates the trajectory of $AV_T$ such that it collides with any vehicle or object other than $AV_{adv}$. A simple example of suitable adversarial reward function is $R_t = \eta_T\times C_T - d(AV_T, AV_{adv}) - \eta_{adv}\times C_{adv}$, where $\eta_x$ determines whether $AV_{x\in \{T, adv\}}$ has collided with another object, $C_T > 0$ is the value of $AV_T$ being in a collision, and $C_{adv} > 0$ captures how undesired direct collisions are for $AV_{adv}$.
	\item \emph{Trajectory manipulation}: The aim of this objective is to exploit collision avoidance maneuvers \cite{pierpaoli2015altering} of $AV_T$ to shift its natural trajectory towards an arbitrary alternative. An instance of corresponding reward function is $R_t = -d(AV_T, AV_{adv}) - d(AV_T, S'(t)))$, where $S'(t)$ is the position of $AV_T$ at time $t$ if it were following the desired adversarial trajectory $S'$.
\end{enumerate}

Our proposed framework allows for arbitrary modification and definition of reward functions corresponding to any one of the aforementioned objectives, as long as the reward functions satisfy the fundamental requirements of rationality \cite{bowling2001rational}. It is noteworthy that not all objectives are necessarily reachable, and this framework provides a best-effort solution which is guaranteed to converge if the objective is reachable.

\subsection{Deep RL Model}
While there are numerous deep RL algorithms with proven success \cite{arulkumaran2017brief}, not all are suitable for learning autonomous navigation policies. Driving falls within the domain of continuous control, meaning that the actuation parameters, such as acceleration, brake, and steering, can take any value from a bounded range of real numbers. Many of the algorithms proposed for deep RL, like Deep Q-Networks (DQNs) \cite{arulkumaran2017brief}, are designed for problems with discrete and low-dimensional action spaces. It is of course possible to discretize sets of continuous values, but this would lead to explosion of dimensions, thus rendering discrete approaches such as DQN infeasible. This limitation has motivated various efforts towards development of continuous deep RL algorithms, resulting in recent proposal of techniques such as Deep Deterministic Policy Gradient (DDPG), variants of Trust Region Policy Optimization, and variants of Asynchronous Advantage Actor Critic (A3C) \cite{arulkumaran2017brief}. 

Since the basis of deep RL is function approximation with deep neural networks, implementation of such algorithms requires suitable choices (i.e., tuning) of the model architecture and its hyper-parameters. A common approach to this problem is to begin with an architecture and parameters from successful implementations of similar problems. For the purposes of this work, some options include those presented in \cite{lillicrap2015continuous}. An important consideration in choosing or tuning architectures is the type of inputs. Conventionally, if the state observations of the RL agent are pixel values of the environment, convolutional architectures are considered. If states are numerical vectors, multi-layer perceptrons may prove most feasible, and if states are sequential data (e.g., time-domain samples), a recurrent architecture such as Long Short-Term Memory (LSTM) provides a good starting point for model tuning. 

Another important aspect of deep RL models is the choice of exploration mechanism. Multiple studies (e.g., \cite{fortunato2017noisy}) report that the classical $\epsilon$-greedy mechanism fails to provide efficient performance in continuous and multi-action scenarios. Alternative approaches include parameter-space noise methods, such as NoisyNet \cite{fortunato2017noisy} and the Ornstein-Uhlenbeck process \cite{lillicrap2015continuous}.

To preserve the generality and flexibility of our solution, this framework allows for arbitrary choice of model type, architecture, and parameters. Of course, it is noteworthy that due to the stochastic nature of deep RL and exploration processes, different models may yield different results. Hence, to obtain valid comparisons, it is essential to use the same model architecture and configuration across the domain of comparison.

\subsection{Metrics}
We propose three classes of metrics for quantitative benchmarking of experiments: those obtained from the training process, those measured at test-time, and numerable parameters of initial environment configurations. One instance of training-time metrics is the number of iterations (alternatively, steps or episodes) to reach the threshold of convergence. Lower values of this measurement, averaged over repeated experiments, indicates that finding a strategy to jeopardize the safety of algorithm under test requires fewer explorations into multi-dimensional observations, and may signal a weaker resilience to the simulated type of hazard. Examples of test-time metrics include time or distance to incur damage, for which the lower values indicate weaker robustness of the algorithm. Another test-time metric is levels of damage incurred to each AV: higher damage to $AV_T$ indicates higher safety risks, while higher damage to $AV_{adv}$ demonstrates the higher economic cost of intentional attacks, which in turn indicates lower safety risks for $AV_T$. The initial environment configuration may include the number, density, and initial arrangement of AVs that may impact the reliable behavior of the system. For instance, similar to the case of TCAS, the collision avoidance algorithm employed by an $AV_T$ may fail to retain its safety criteria when the number of other vehicles in the environment is larger than a certain threshold. In this instance, the maximum number of agents that can safely coexist with $AV_T$ can be noted as a measure of capacitive threshold. 

Also, it is indeed noteworthy that due to the stochastic nature of deep RL and some simulation platforms, repeated measurements are necessary to obtain valid results.

\section{Case Study}\label{sec-case}
To demonstrate the application of our framework in practice, we have compared the resilience of two ground AVs to adversarial agents trained for direct collisions. One agent, $AV_D$, is an AV equipped with motion planning and collision avoidance policies obtained from the deep RL approach proposed in \cite{lillicrap2015continuous} \footnote{The only notable difference between the model of \cite{lillicrap2015continuous} and this implementation is that the state vector of the latter incorporates a vector of distances with opponents, in addition to the state parameters considered in the former. This alteration is due to the need for accurate positioning of other agents in the collision avoidance policy}. The other agent, $AV_M$, is the Olethros driver bot \cite{loiacono2013simulated} that adopts a model-predictive control approach towards navigation and collision avoidance. Table \ref{tab-config} describes the configuration of main components for this experiment. We have chosen TORCS as the simulation platform of this experiment, as it satisfies the necessary requirements noted in Section \ref{sec-architecture}. A further advantage of TORCS is the availability of an OpenAI Gym interface to its environment, thus making its integration with the TensorFlow deep learning platform easier. We configured the environment to include multiple AVs, which include $AV_T$ and $AV_{adv}$, as well as a group of arbitrarily selected autonomous bots from the defaults of TORCS. 

\begin{table}
	{\renewcommand{\arraystretch}{2}%
		\caption{Experiment Settings}
		\label{tab-config}
		\setlength{\tabcolsep}{3pt}
		\begin{tabular}{|p{75pt}|p{150pt}|}
			\hline
			Component& 
			Configuration\\
			\hline
			Environment& 
			TORCS, multi-agent\\
			Objective& 
			Direct collision\\
			Reward function& 
			$R_t = \eta\times 200 - d(AV_T, AV_{adv})$\\
			Deep RL model& 
			DDPG - same architecture as \cite{lillicrap2015continuous}\\
			Exploration mechanism& 
			Ornstein-Uhlenbeck\\ 
			Training-time metrics& 
			Number of episodes to convergence\\
			Test-time metrics& 
			Number of (simulated) seconds to collision\\
			\hline
	\end{tabular}}
	\label{tab0}
\end{table}


Table~\ref{tab-results} presents the training progress of adversarial agent for both $AV_M$ and $AV_D$. It can be seen that $AV_M$ allows faster convergence (473 vs 889 episodes), and has a lower return value than $AV_D$. It is noteworthy than higher values of minimum time to collision indicate greater resilience, while higher values of optimal return represent weaker robustness. Accordingly, these observations indicate that $AV_M$ is more predictable and less complex than $AV_D$, while $AV_D$ is more resilient but less robust compared to $AV_M$. Similarly, test-time results presented in Table~\ref{tab-results} indicate that $AV_M$ allows collisions at a much sooner time than $AV_D$. Thus, we can conclude that in the selected environment, the motion planning and collision avoidance algorithm of $AV_D$ is more resilient to direct collisions than that of $AV_M$.
\begin{table}
	\centering
	{\renewcommand{\arraystretch}{2}%
		\caption{Experiment Results - Averaged over 100 runs}
		\label{tab-results}
		\setlength{\tabcolsep}{3pt}
		\begin{tabular}{|p{100pt}|c|c|}
			\hline
			Metric&
			$AV_M$&
			$AV_D$\\ 
			\hline
			Number of episodes to convergence&
			470& 
			890\\
			Optimal return& 
			13900&
			15400\\
			Time to collision& 
			22.44s&
			51.31s\\
			\hline
	\end{tabular}}
	\label{tab3}
\end{table}

\section{Summary and Future Directions}\label{sec-conclusion}
We have proposed a process flow and framework that utilize adversarial deep reinforcement learning to measure the reliability of motion planning and collision avoidance mechanisms in autonomous vehicles. We have established the advantages of this framework over current benchmarking schemes, which include flexibility and generality, adaptive probing by training adversarial policies against the particular system-under-test, sample-efficient and customizable exploration mechanisms, and provision of baseline (i.e., worst-case) measurements for benchmarking and comparison among heterogeneous systems.

The straightforward architecture of the proposed framework presents a number of potential venues for further research. An immediate next step is to apply this framework to prominent and recently published techniques for motion planning and collision avoidance, with the aim of creating reference benchmarks for use in relevant research projects. Another promising venue is to check the applicability of recent publications (e.g.,\cite{mandlekar2017adversarially}) that claim training under adversarial perturbations can enhance the resilience and robustness of the policy. Hence, a potential mitigation and defense technique may emerge from investigating the training of adversarial policy and reinforcement learning models of collision avoidance in conjunction.

\fussy
%
\end{document}